\documentclass[letterpaper]{article} 
\usepackage{aaai2026}  
\usepackage{times}  
\usepackage{helvet}  
\usepackage{courier}  
\usepackage[hyphens]{url}  
\usepackage{amsmath}
\usepackage{graphicx} 
\usepackage{natbib}  
\usepackage{caption} 
\usepackage{algorithm}
\usepackage{algorithmic}
\usepackage{booktabs}
\usepackage{newfloat}
\usepackage{listings}
\DeclareCaptionStyle{ruled}{labelfont=normalfont,labelsep=colon,strut=off} 
\floatstyle{ruled}
\newfloat{listing}{tb}{lst}{}
\floatname{listing}{Listing}
\title{Coherent Multi-Agent Trajectory Forecasting in Team Sports with CausalTraj}
\author {
    Wei Zhen Teoh
}
\affiliations{
    wzteoh.works@gmail.com
}

\begin{document}

\maketitle

\begin{abstract}
Jointly forecasting trajectories of multiple interacting agents is a core challenge in sports analytics and other domains involving complex group dynamics. Accurate prediction enables realistic simulation and strategic understanding of gameplay evolution. Most existing models are evaluated solely on per-agent accuracy metrics (\textit{minADE}, \textit{minFDE}), which assess each agent independently on its best-of-\textit{k} prediction. However these metrics overlook whether the model learns which predicted trajectories can jointly form a plausible multi-agent future. Many state-of-the-art models are designed and optimized primarily based on these metrics. As a result, they may underperform on joint predictions and also fail to generate coherent, interpretable multi-agent scenarios in team sports. We propose CausalTraj, a temporally causal, likelihood-based model that is built to generate jointly probable multi-agent trajectory forecasts.
To better assess collective modeling capability, we emphasize joint metrics (\textit{minJADE}, \textit{minJFDE}) that measure joint accuracy across agents within the best generated scenario sample.
Evaluated on the NBA SportVU, Basketball-U, and Football-U datasets, CausalTraj achieves competitive per-agent accuracy and the best recorded results on joint metrics, while yielding qualitatively coherent and realistic gameplay evolutions.
\end{abstract}

\begin{links}
    \link{Code}{https://github.com/wezteoh/causaltraj}
    \link{Project Page}{https://causaltraj.github.io}
\end{links}

\section{Introduction}

Artificial intelligence is increasingly applied in sports analytics, enabling automatic understanding of tactical patterns, player roles, and decision-making dynamics~\cite{tuyls2021game,wang2024tacticai,wu2022sports}. 
A key foundation for such analysis is \textit{trajectory prediction} -- forecasting the future motion of players and the ball from historical observations. 
Accurate trajectory forecasting not only aids tactical interpretation but also enables realistic simulation and generation of gameplay sequences. 
Beyond sports, this problem also appears in autonomous driving~\cite{8793868} and crowd navigation~\cite{Lisotto_2019_ICCV}, where interaction and coordination among multiple agents are central. Trajectory forecasting in these domains is inherently challenging: the future is stochastic and multimodal, and each agent’s motion depends on the collective configuration of all others.

Most trajectory forecasting models are evaluated solely using per-agent metrics (\textit{minADE} and \textit{minFDE}), which score each agent independently based on its best-of-k predicted trajectories. Many state-of-the-art works on sports datasets \cite{mao2023leapfrog, lee2024mart, fu2025moflowonestepflowmatching} design their models and training objectives around these metrics. For instance, their losses supervise multi-hypothesis trajectory selection independently for each agent, without modeling which predicted trajectories across agents should fit together to form a joint multi-agent future. As a result, models may achieve good per-agent scores yet still underperform on joint prediction. They may also generate trajectories that look reasonable individually but fail to form coherent~\footnote{We use "coherent" to refer to a property of multi-agent trajectories in which agents' relative motions are physically plausible and mutually compatible, e.g. ball motion aligning with player motion and players positioning themselves in coordinated formations.} multi-agent evolutions. The importance of assessing joint predictions in trajectory forecasting has formerly been highlighted in other domains involving group dynamics~\cite{weng2023jointmetricsmatterbetter}. In team sports, the state of play is defined by joint behaviors. As we cannot assume oracle knowledge of groundtruth to select compatible marginal predictions at test time, reliable joint predictions are fundamental.

We view the capability to learn the true joint distribution as the central goal to pursue. When the true joint distribution is captured, scenario-level coherence and strong per-agent accuracy should emerge naturally as byproducts rather than competing objectives.
Inspired by recent successes of causal (autoregressive) architectures in language (LLMs) and 3D environment generation ~\cite{radford2019language,parkerholder2024genie2}, we revisit the temporally causal framework for trajectory prediction.
Modeling causality in time allows spatial and inter-agent dynamics to evolve step-by-step, rather than being compressed into a fixed global latent representation. Combined with likelihood-based objectives, it captures multimodal transition uncertainty while providing high capacity for joint modeling.

We thus propose \textbf{CausalTraj}, a temporally causal, likelihood-based model to generate multi-agent trajectories.

Our contributions are summarized as follows:
\begin{itemize}
\item We present a causal model that combines established components of spatiotemporal modeling and multimodal likelihood prediction into an effective design for joint multi-agent trajectory forecasting. We further introduce lightweight adaptations to enhance spatial information and enable efficient causal training.
\item We highlight the importance of evaluating joint trajectory modelling capability through joint metrics (\textit{minJADE}, \textit{minJFDE}), which capture aspects of collective behavioral structure that per-agent metrics overlook. Our qualitative visualizations suggest that improvements in joint metrics align with models' ability to generate perceptually more coherent gameplay outcomes.
\item Through experiments on established benchmarks, we show that our models achieve competitive per-agent accuracy, the best recorded joint-metric performance, and generate qualitatively coherent gameplay evolutions, supporting their potential for applications such as gameplay simulation and tactical analysis.
\end{itemize}

\section{Related Work}

\paragraph{Stochastic Multimodality.} Multimodality in trajectory prediction has been addressed through various generative paradigms. Early methods explicitly parameterize output distributions~\cite{IvanovicP19}, while latent-variable approaches such as VAEs~\cite{zhan2019generating,xu2022socialvae,xu2025sportstraj,xu2022GroupNet} capture stochasticity via latent codes. More recent diffusion- and flow-based models~\cite{mao2023leapfrog,bae2024singulartrajectory,fu2025moflowonestepflowmatching} rely on distribution transformation principles to achieve state-of-the-art per-agent accuracy.

\paragraph{Spatiotemporal Relation Modeling.}
Spatial interactions are commonly modeled using graph and hypergraph formulations~\cite{li2020evolvegraph,xu2022GroupNet,xu2023eqmotionequivariantmultiagentmotion} or transformer-based attention mechanisms~\cite{NIPS2017_3f5ee243,yuan2021agent,fu2025moflowonestepflowmatching}. For temporal dependencies, earlier works employed RNNs~\cite{10.1162/neco.1997.9.8.1735,zhan2019generating,hauri2021multi}, later replaced by transformers~\cite{DBLP:journals/corr/abs-2003-08111,yuan2021agent}. More recently, structured state-space models such as the Mamba family~\cite{mamba,mamba2} combine compact state representations with attention-like contextual modeling, offering an efficient alternative~\cite{huang2025trajectorymambaefficientattentionmamba,xu2025sportstraj}.

\paragraph{Output Structure.}
Causal (autoregressive) formulations have shown success in earlier works~\cite{IvanovicP19,zhan2019generating}.
However, many recent state-of-the-art methods achieving top per-agent metrics on sports datasets adopt designs that predict the full future trajectory horizon in parallel~\cite{mao2023leapfrog,lee2024mart,fu2025moflowonestepflowmatching}, often via an intermediate global latent representation.
This formulation, paired with multi-sample output heads, enables control over prediction diversity and simplifies optimization for marginal (per-agent) accuracy. However, for joint trajectory predictions, predicting all agents and timesteps simultaneously require the outputs across timesteps to be conditionally independent given the global latent representation. As a result, modeling interdependent agent dynamics over a long horizon would probably require a huge and expressive latent state.

\paragraph{Evaluation.} \textit{minADE} and \textit{minFDE} are standard metrics for trajectory prediction, measuring average and final positional errors independently for each agent. Recent work~\cite{weng2023jointmetricsmatterbetter} emphasized the importance of joint metrics including \textit{minJADE} and \textit{minJFDE} (where J stands for "joint") for evaluating collective modelling capability and demonstrated it on pedestrian datasets. Our study extends this perspective to the sports domain.

\paragraph{Our Approach.}
We revisit the causal formulation, motivated by its proven success in capturing high-order dependencies across sequential domains such as language and 3D environment generation~\cite{radford2019language,parkerholder2024genie2}.
By modeling interactions step-by-step instead of compressing them into a latent state, the causal structure reduces the requirement for latent capacity. We directly parameterize multimodal trajectory likelihoods in the output space, enabling exact optimization of probabilistic objectives without approximate inference.
Our architecture combines a transformer-based inter-agent relation encoder with spatial adaptations and a Mamba2-based temporal encoder under a unified causal framework.
By integrating these complementary components, CausalTraj produces coherent multi-agent forecasts that excel on joint metrics.

\section{Problem Formulation}

We consider $N$ interacting agents in a 2D coordinate space. Each agent $i$ has an observed historical trajectory 
\[
X_{i,1:P} = [x_{i,1}, x_{i,2}, \ldots, x_{i,P}] \in \mathbf{R}^{P \times 2},
\]
where $x_{i,t}$ denotes the spatial position of agent $i$ at time $t$.  
Given the historical trajectories of all agents,
\[
X_{1:P} = [X_{1,1:P}, X_{2,1:P}, \ldots, X_{N,1:P}] \in \mathbf{R}^{N \times P \times 2},
\]
our goal is to forecast their joint future trajectories
\[
\hat{X}_{P+1:T} = [\hat{X}_{1,P+1:T}, \hat{X}_{2,P+1:T}, \ldots, \hat{X}_{N,P+1:T}],
\]
where $\hat{X}_{i,P+1:T} \in \mathbf{R}^{F \times 2}$ represents the predicted future path of agent $i$ for $F = T - P$ timesteps.

A single joint prediction $\hat{X}_{P+1:T}$ represents one possible future configuration of all agents, which we refer to as a \textbf{scenario}.  
The objective of multi-agent trajectory forecasting is thus to learn a model that estimates the conditional distribution
\[
p(X_{P+1:T} \mid X_{1:P}),
\]
allowing sampling of diverse and probable joint future scenarios.

\section{Causal Likelihood Modeling Framework}
We model the conditional distribution $p(X_{ P+1:T} \mid X_{1:P})$ as a product of causal likelihoods over timesteps:
\[
p(X_{P+1:T} \mid X_{1:P}) = \prod_{t=P}^{T-1} p(X_{t+1} \mid X_{1:t}).
\]
To make learning more stable and focus on motion dynamics, our model predicts the per-timestep displacement of all agents,
\[
\Delta X_{t+1} = X_{t+1} - X_t,
\]
and models the corresponding conditional distribution $p(\Delta X_{t+1} \mid X_{1:t})$.

\paragraph{Autoregressive sampling and parallel training.}
During inference, the model samples displacements $\hat{\Delta X}_{t+1}$ from the predicted distribution and updates positions recursively as $\hat{X}_{t+1} = \hat{X}_t + \hat{\Delta X}_{t+1}$. 

We train the model in parallel across time-steps in teacher forcing fashion. The model parameters are optimized to maximize the likelihood of the groundtruth positions at the next timestep.

\paragraph{Mixture-of-Gaussians output and training objective.}
To capture the multimodal nature of gameplay evolution, we model the per-step displacement with a mixture of $M$ Gaussians:
$p(\Delta X_{t+1}\mid X_{1:t}) = \sum_{m=1}^{M} \pi_{t+1,m} \, \mathcal{N}(\Delta X_{t+1}; \mu_{t+1,m}, \Sigma_{t+1,m})$.
At each timestep, the network outputs:
\begin{itemize}
    \item $M$ mixture logits (converted to weights $\pi$ by softmax),
    \item $M \times N \times 2$ means $\mu$ (per agent and component), and
    \item $M \times N \times 3$ Cholesky parameters for block-diagonal covariances.
\end{itemize}
For tractability, we assume conditional independence across agents within each component within a timestep, so $\Sigma_{t,m}=\mathrm{blockdiag}(\Sigma_{t,m,1},\ldots,\Sigma_{t,m,N})$ with each $\Sigma_{t,m,n}\in\mathbf{R}^{2\times2}$ parameterized via a lower-triangular Cholesky factor $L_{t,m,n}$. Although within each component we assume zero cross-agent covariance, the mixture can still represent interdependent joint structure because the shared mixture weights already couple agents’ outcomes.

Training maximizes the likelihood of the ground-truth displacements 
$Y_t = \Delta X_{t}$
under the predicted mixture distribution.  
We minimize the negative log-likelihood (NLL), 
augmented with an entropy regularizer on the mixture weights 
to discourage component collapse early in training. 
The loss for each multi-agent trajectory sample is defined as:

{\small
\[
\mathcal{L}_{\text{NLL}}
= -\,\mathbf{E}_{t}\!
\left[
\log\!\Bigg(
\sum_{m=1}^{M} 
\hat{\pi}_{t,m}\,
\mathcal{N}\!\big(Y_{t};\,\hat{\mu}_{t,m},\,\hat{\Sigma}_{t,m}\big)
\Bigg)
\right],
\]
}
\[
\mathcal{L}_{\text{ent}}
= -\,\frac{1}{\log M}
\sum_{m=1}^{M}
\hat{\pi}_{t,m}\,
\log\!\big(\hat{\pi}_{t,m}+\varepsilon\big),
\]
\[
\mathcal{L}
= \mathcal{L}_{\text{NLL}}
- \lambda_{\text{ent}}\,
\mathcal{L}_{\text{ent}}.
\]

We use $M=8$ and $\lambda_{\text{ent}}{=}0.05$ in all experiments.

\paragraph{Velocity augmentation.}
We include instantaneous velocity features, computed as current timestep displacement from previous, as auxiliary inputs alongside absolute positions.

More training, loss derivation and model implementation details can be found in appendix~\ref{appendix:training} and our open source code~\footnote{\url{https://github.com/wezteoh/causaltraj}}.

\section{Model Architecture}

\begin{figure}[!t]
\centering
\includegraphics[width=0.95\columnwidth]{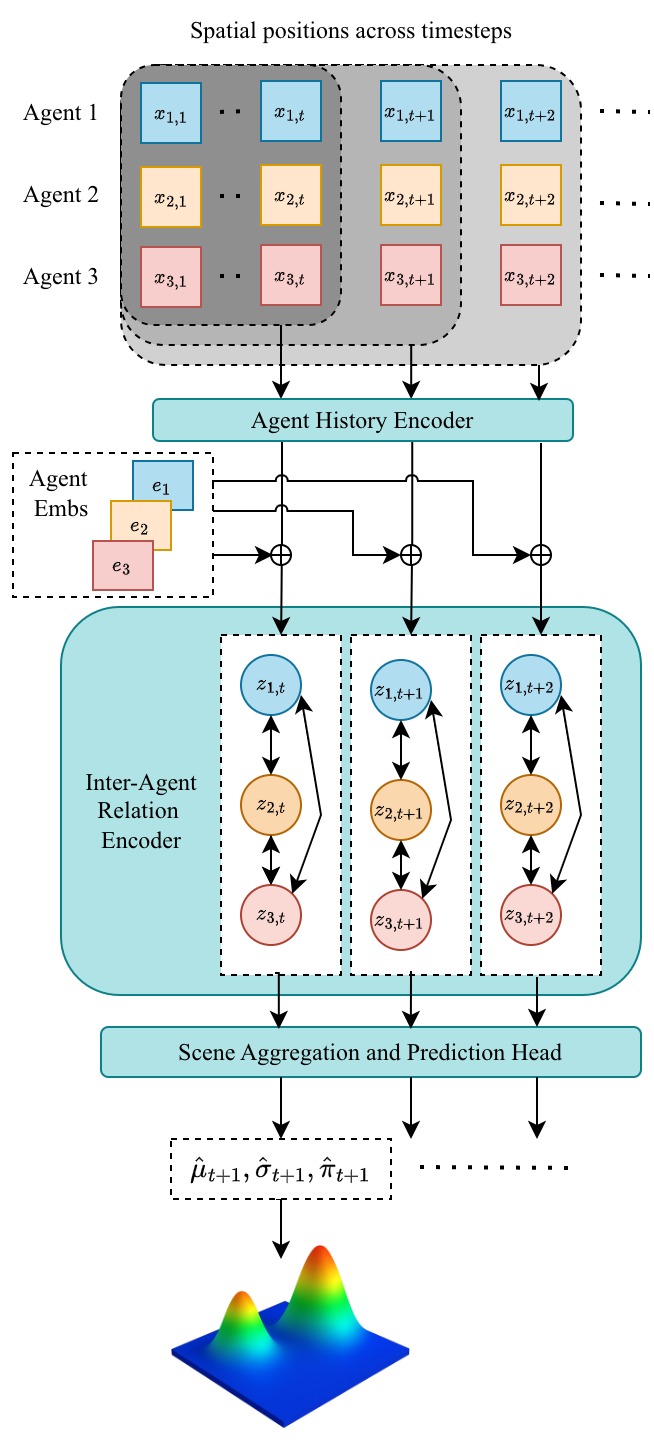} 
\caption{CausalTraj Model Overview}
\label{fig1}
\end{figure}

The high level architecture is illustrated in Figure~\ref{fig1}.

\subsection{Agent History Encoder}
At this stage, each agent’s historical trajectory information is encoded independently, without inter-agent interactions. The output at this stage is a sequence of agent features $Z$, where $z_{i,t}$ is time-accumulated agent feature up to time $t$ for agent $i$. We experimented with two encoder module designs.

\paragraph{Causal PointNet Encoder.}
We follow the
previous approach~\cite{fu2025moflowonestepflowmatching}
to adopt a PointNet-style encoder~\cite{qi2016pointnet}.
Unlike the original implementation, which applies global max-pooling
across all timesteps, we adapted it into a causal formulation. we introduce a
\textit{lookback max-pooling} operation that, for each timestep $t$, aggregates features only
for timesteps $t'\!\le\!t$, implemented via zero-padding and sliding-window
pooling. The feedforward sequence is as follow:
{\small
\[
\text{MLP}
\!\rightarrow\!
\text{Lb-MaxPool}
\!\rightarrow\!
\text{Concat}
\!\rightarrow\!
\text{MLP}
\!\rightarrow\!
\text{Lb-MaxPool}
\!\rightarrow\!
\text{MLP}
\]
}
The pooled context is concatenated with previous MLP (multi-layer perceptron) outputs, timestep-wise, at the concat layer. 
Our adaptation preserves temporal causality,
enables efficient teacher-forced parallel training,
and retains the hierarchical feature aggregation benefits of PointNet.

\paragraph{Mamba2 Encoder.}
In our model, we experiment swapping the Causal PointNet encoder for the Mamba2 module. We added 2-layer MLP to project the initial trajectory information per timestep to higher dimension. Each agent's feature sequence is passed through the Mamba2 layers independently. The sequence of hidden states output by Mamba2 serves as the compressed representations of each agent's trajectory up to the corresponding timesteps. This configuration yields improved performance on several benchmark datasets.

\subsection{Agent Embedding} We update the encodings $Z$ by concatenating them with learned agent embeddings. There are only 3 different embedding vectors: 2 teams and the ball. The concatenated features are passed through another MLP layer.

\subsection{Inter-Agent Relation Encoder}
Given stacked agent encodings $Z_t\in\mathbf{R}^{N\times d}$ (agent dim, encoding dim) at timestep $t$, we apply $N$ transformer-style encoder blocks to capture inter-agent interactions. Each block consists of (i) inter-agent self-attention computed independently at each $t$, followed by (ii) an agentwise feedforward layer. No cross-temporal attention is used here.

While standard self-attention can implicitly model interactions via content similarity, it does not explicitly encode pairwise spatial geometry (e.g., exact Euclidean displacements). For multi-agent motion, such geometry is often predictive. We therefore introduce additional N blocks of \textit{Spatial Relation Transformer Encoder} that augments attention with a learned function of pairwise offsets.

At each timestep $t$, let $X_t\in\mathbf{R}^{N\times 4}$ denote agents' current positions and velocities. We construct a pairwise “mesh” tensor $M_t \in \mathbf{R}^{N \times N \times (2d+4)}$, where
\[
M_t[q,k] \;=\; \big[\,x_{q,t}-x_{k,t}\;;\;z_{q,t}\;;\;z_{k,t}\,\big] \in \mathbf{R}^{d_\text{mesh}},
\]
$M_t[q,k]$ is a concatenated vector where $q$ indexes the query agent and $k$ the key/value agent. Each query is projected from a $z_{q,t}$, and the corresponding keys/values are projected from the vectors in the row $q$ of $M_t$; multi-head scaled dot-product attention is then applied over the key agent dimension. This design exposes exact relative displacement to be used to compute value features, prior to weighted aggregation via attention. We note that the above adaptation to embed pairwise spatial information into transformer shares some similarities with a prior work~\cite{10.1007/978-3-030-58610-2_30}.

\subsection{Scene Aggregation and Prediction Head}
We concatenate the learned latent features for each agent with their corresponding positional and velocity information, at respective timestep. The resulting feature vectors are then compressed in dimensionality through an agentwise MLP (1-layer). Next, we aggregate all agents’ features within the same timestep by concatenation, forming a single scene-level representation. This representation is passed through 3-layer MLP to produce the parameters of a Mixture of Gaussians, which defines the predicted distribution of the agents’ collective displacement deltas for the next timestep.

\section{Experiments}
\subsection{Datasets}
We evaluate our model on three multi-agent sports trajectory datasets: NBA SportVU, Basketball-U, and Football-U.
NBA SportVU built from NBA player movement logs \footnote{\url{https://github.com/linouk23/NBA-Player-Movements}} has been widely used as a benchmark for state-of-the-art trajectory prediction. Each sequence contains 30 frames recorded at 5 Hz, capturing 10 players and a ball. Following prior work \cite{fu2025moflowonestepflowmatching, mao2023leapfrog}, we use the first 10 frames as context and predict the next 20.
Basketball-U \cite{xu2025sportstraj}, derived from the NBA dataset \cite{zhan2019generating}, consists of 50-frame sequences. Football-U \cite{xu2025sportstraj}, built from the NFL Big Data Bowl dataset\footnote{\url{https://github.com/nfl-football-ops/Big-Data-Bowl}}
, contains 50-frame samples at 10 Hz, featuring 22 players and a ball. Originally, Basketball-U and Football-U test splits contained subsets to be used for imputation tasks with dedicated masks; we instead use all data for full future prediction of the final 20 frames.
For NBA datasets, we follow the convention in the previous works to scale all results by $28/94$ to convert foot units to metres \footnote{conversion is slightly off, but we stick to them to maintain result comparability against previous works.}.

\begin{table*}[!htbp]
\centering
\small
\setlength{\tabcolsep}{6pt}
\begin{tabular}{lcccccc}
\toprule
\textbf{Time} &
\shortstack[c]{\textbf{GroupNet}\\CVPR'22} &
\shortstack[c]{\textbf{LED}\\CVPR'23} &
\shortstack[c]{\textbf{MoFlow}\\\textbf{(joint obj.)}} &
\shortstack[c]{\textbf{MoFlow}\\CVPR'25} &
\shortstack[c]{\textbf{CausalTraj}\\\textbf{(C-PointNet)}} &
\shortstack[c]{\textbf{CausalTraj}\\\textbf{(Mamba2)}} \\

\midrule
1.0s & 0.25/0.32 & 0.21/0.27 & 0.28/0.39 & 0.18/0.25 & 0.15/0.21 & \textbf{0.14}/\textbf{0.20} \\
2.0s & 0.47/0.68 & 0.44/0.56 & 0.48/0.71 & 0.34/\textbf{0.47} & 0.34/0.50 & \textbf{0.33}/0.49 \\
3.0s & 0.71/0.99 & 0.69/0.84 & 0.68/1.01 & \textbf{0.52}/\textbf{0.67} & 0.55/0.78 & 0.54/0.78 \\
4.0s & 0.95/1.22 & 0.81/1.10 & 0.89/1.32 & \textbf{0.71}/\textbf{0.87} & 0.77/1.01 & 0.77/1.02 \\
\midrule
1.0s & 0.50/0.77 & 0.34/0.64 & 0.40/0.67 & 0.37/0.68 & 0.28/0.50 & \textbf{0.27}/\textbf{0.49} \\
2.0s & 1.04/1.91 & 0.78/1.55 & 0.81/1.61 & 0.80/1.61 & \textbf{0.62}/\textbf{1.18} & \textbf{0.62}/1.21 \\
3.0s & 1.61/2.98 & 1.22/2.36 & 1.27/2.55 & 1.25/2.49 & \textbf{0.98}/\textbf{1.86} & 1.00/1.93 \\
4.0s & 2.12/3.72 & 1.63/2.99 & 1.72/3.33 & 1.69/3.31 & \textbf{1.34}/\textbf{2.47} & 1.38/2.57 \\
\bottomrule
\end{tabular}
\caption{Performance on SportVU NBA dataset. Each cell shows \textit{minADE$_{20}$/minFDE$_{20}$} in metre unit in the upper block and \textit{minJADE$_{20}$/minJFDE$_{20}$} in metre unit in the lower block. Best results are bolded.}
\label{tab:nba_metrics}
\end{table*}

\begin{table*}[!htbp]
\centering
\small
\setlength{\tabcolsep}{5pt}
\begin{tabular}{llccccc}
\toprule
\textbf{Dataset} &
\shortstack[c]{\textbf{Frame} \\ \textbf{count}} &
\shortstack[c]{\textbf{Sports-Traj}\\ ICLR'25} &
\shortstack[c]{\textbf{MoFlow}\\\textbf{(joint obj.)}} &
\shortstack[c]{\textbf{MoFlow}\\ CVPR'25} &
\shortstack[c]{\textbf{CausalTraj}\\ \textbf{(C-PointNet)}} &
\shortstack[c]{\textbf{CausalTraj}\\\textbf{(Mamba2)}} \\
\midrule
\textbf{Basketball-U}
 & 10 & 0.84/1.50 & 0.33/0.50 & \textbf{0.24/0.32} & 0.25/0.37 & \textbf{0.24}/0.34 \\
 & 20 & 1.52/2.61 & 0.63/0.91 & \textbf{0.50/0.61} & 0.58/0.74 & 0.56/0.71 \\
\cmidrule(lr){2-7}
 & 10 & 0.85/1.51 & 0.57/1.15 & 0.56/1.11 & 0.46/0.87 & \textbf{0.45/0.85} \\
 & 20 & 1.52/2.62 & 1.21/2.34 & 1.18/2.30 & 0.98/\textbf{1.77} & \textbf{0.97/1.77} \\
\midrule
\textbf{Football-U}
 & 10 & 3.38/2.89 & 0.29/0.61 & \textbf{0.16}/\textbf{0.27} & 0.19/0.37 & \textbf{0.16}/0.31 \\
 & 20 & 3.64/3.33 & 0.76/1.62 & \textbf{0.42}/\textbf{0.80} & 0.57/1.15 & 0.50/0.99 \\
\cmidrule(lr){2-7}
 & 10 & 3.40/2.98 & 0.42/0.94 & 0.40/0.92 & 0.41/0.91 & \textbf{0.37/0.85} \\
 & 20 & 3.66/3.46 & 1.19/2.87 & 1.16/2.82 & 1.17/2.74 & \textbf{1.12/2.68} \\
\bottomrule
\end{tabular}
\caption{Performance on the Basketball-U (metres) and Football-U (yards) datasets. Each cell shows \textit{minADE$_{20}$/minFDE$_{20}$} in the upper half and \textit{minJADE$_{20}$/minJFDE$_{20}$} in the lower half, evaluated over 10- and 20-frame horizons. Lower is better.}
\label{tab:combined_metrics}
\end{table*}

\subsection{Metrics}
We evaluate models using four standard metrics: \textit{minADE\textsubscript{k}}, \textit{minFDE\textsubscript{k}}, \textit{minJADE\textsubscript{k}}, and \textit{minJFDE\textsubscript{k}}.

\paragraph{Per-agent metrics.}
For each agent $i$, given $k$ predicted future trajectories $\{\hat{X}^j_{i,P+1:T}\}_{j=1}^k$ and the ground-truth trajectory $X_{i,P+1:T}$, we define:

\begin{equation}
\textit{minADE}_k = \frac{1}{N} \sum_{i=1}^N \min_j 
\frac{1}{F} \sum_{t=P+1}^{T} 
\lVert \hat{x}^j_{i,t} - x_{i,t} \rVert
\end{equation}

\begin{equation}
\textit{minFDE}_k = \frac{1}{N} \sum_{i=1}^N 
\min_j \lVert \hat{x}^j_{i,T} - x_{i,T}\rVert
\end{equation}

Intuitively, \textit{minADE} (minimum average displacement error) and \textit{minFDE} (minimum final displacement error) measure the average and final positional errors of the most accurate predicted trajectory for each agent, selected from $k$ generated candidates. However, these metrics evaluate agents independently, i.e. each agent’s best trajectory may come from a different predicted scenario by the model. Due to the nature of these metrics, in many existing methods, trajectories for different agents are even produced as independent marginal samples, without the notion of scenario grouping. This limits their ability to produce realistic simulations without oracle guidance.

\paragraph{Joint metrics.}
To measure joint modelling capability, we compute joint metrics over full multi-agent scenarios  $\{\hat{X}^j_{P+1:T}\}_{j=1}^k$, where each $\hat{X}^j_{P+1:T} = \{\hat{X}^j_{i,P+1:T}\}_{i=1}^N$:

\begin{equation}
\textit{minJADE}_k = 
\min_j \frac{1}{NF} 
\sum_{i=1}^N \sum_{t=P+1}^{T} 
\lVert \hat{x}^j_{i,t} - x_{i,t}\rVert
\end{equation}

\begin{equation}
\textit{minJFDE}_k = 
\min_j \frac{1}{N} 
\sum_{i=1}^N \lVert \hat{x}^j_{i,T} - x_{i,T}\rVert
\end{equation}

These metrics select the best joint prediction among $k$ scenario samples for comparison against the groundtruth. In contrast with the previous metrics, we have to assess a chosen scenario as a whole. Because groundtruth joint trajectories are inherently coherent, achieving low joint error implicitly suggests the model generates joint combinations that are at least coherent configuration-wise. Still, the metrics themselves evaluate joint trajectory accuracy rather than just coherence directly on its own. We use $k=20$ following standard practice.

\begin{figure*}[!htbp]
\centering
\includegraphics[width=0.95\textwidth]{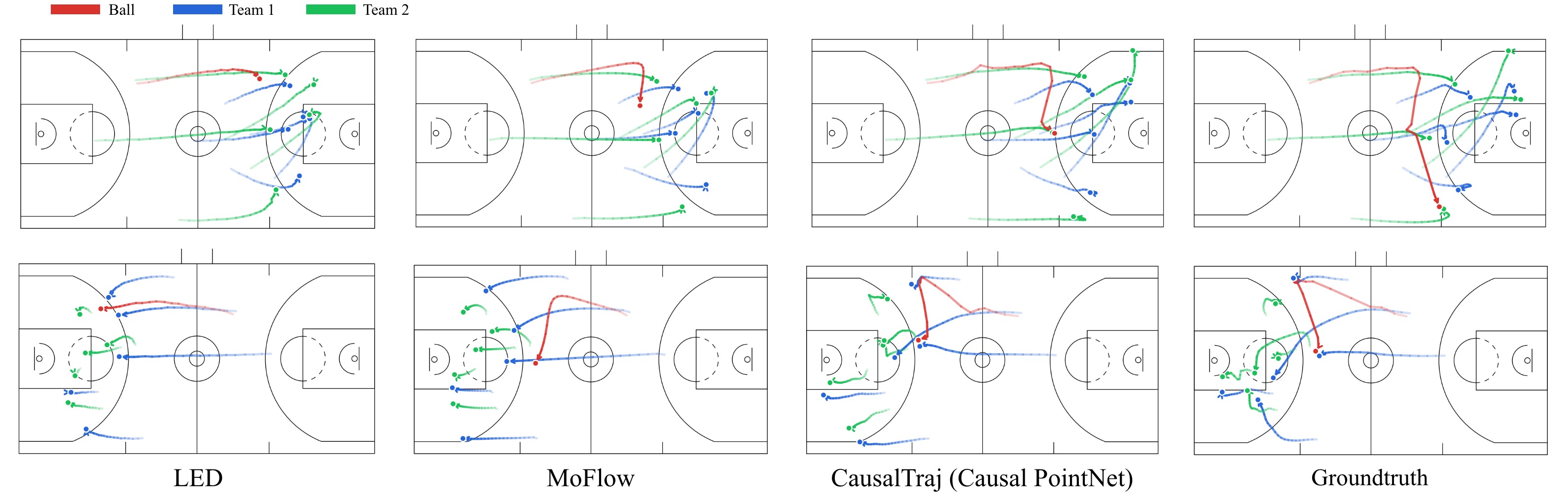} 
\caption{minJADE$_{20}$ sample scenario from selected models vs groundtruth.}
\label{fig:jade}
\end{figure*}

\subsection{Baselines}
For the NBA SportVU dataset, we compare our model against the most recent state-of-the-art approaches on this dataset, including GroupNet \cite{xu2022GroupNet}, LED \cite{mao2023leapfrog}, and MoFlow's denoising models \cite{fu2025moflowonestepflowmatching}. Most of these models primarily focus on marginal predictions, and are optimized for producing diverse per-agent trajectory samples to achieve strong performance on per-agent metrics. For MoFlow and LED, we followed the setup in~\cite{fu2025moflowonestepflowmatching}, grouping the marginal prediction branches of all agents by their branch index to compose scenario samples. Notably, MoFlow also introduced a variant trained with a joint-accuracy objective, which better aligns with our evaluation setting. We reproduce this variant using the official open-source implementation, obtaining slightly improved results compared to the reported numbers. Note that while we would have liked to include more prior works for comparison, many existing models do not provide pretrained weights on this dataset, which prevents consistent re-evaluation under joint metrics. We hope that our study encourages the broader adoption of joint-metric evaluation protocols for multi-agent trajectory prediction in sports analytics.

For the Basketball-U and Football-U datasets, we benchmark against the Sports-Traj model \cite{xu2025sportstraj} which previously recorded the best performance on these datasets. Sports-Traj was trained with a joint objective, aligned with our setting. We also evaluate MoFlow variants trained using their open source implementations on these datasets.

\subsection{Results}

We trained two variants of CausalTraj for each dataset: one with the Causal PointNet Encoder and one with the Mamba2 Encoder. At inference, we draw 20 independent scenario samples for evaluation.

\paragraph{Quantitative Benchmark.} Table~\ref{tab:nba_metrics} reports results on NBA SportVU. On per-agent metrics, MoFlow default version leads in performance for longer horizons (above 2.0s). CausalTraj is slightly better than prior state-of-the-art methods on short horizons (up to 2.0s) and remains competitive on longer horizons, despite not explicitly optimizing for per-agent sample diversity. More importantly, it achieves substantially lower joint metrics (\textit{minJADE}, \textit{minJFDE}), indicating stronger joint modeling capacity.

On Basketball-U and Football-U (Table~\ref{tab:combined_metrics}), we observe similar trends, where MoFlow default version excels in per-agent metrics. CausalTraj excels on short-horizon per-agent metrics and all joint metrics, with the Mamba2 variant standing out. The MoFlow variant trained with a joint-accuracy objective performs almost the same as its default version on the min-based joint metrics across the board. On the other hand, its \textit{averageJADE} (averaged over sampled scenarios rather than taking the best) actually improves substantially. However we did not adopt average metrics in this study as it could favour models predicting single, safe mode. These results also highlight that modeling the true joint multi-agent trajectory distribution is not trivial. Simply modifying the loss function may not suffice without explicitly modeling inter-agent causal dependencies, as done in our approach.

\paragraph{Qualitative Assessment.} 
Figure~\ref{fig:jade} presents qualitative comparisons of the best-performing samples according to \textit{minJADE}. We observe that CausalTraj captures richer and more intricate interactive dynamics among agents such as coordinated directional changes and realistic ball passes between players, whereas prior models tend to produce smoother but less coordinated motions.

\begin{figure*}[!htbp]
\centering
\includegraphics[width=0.95\textwidth]{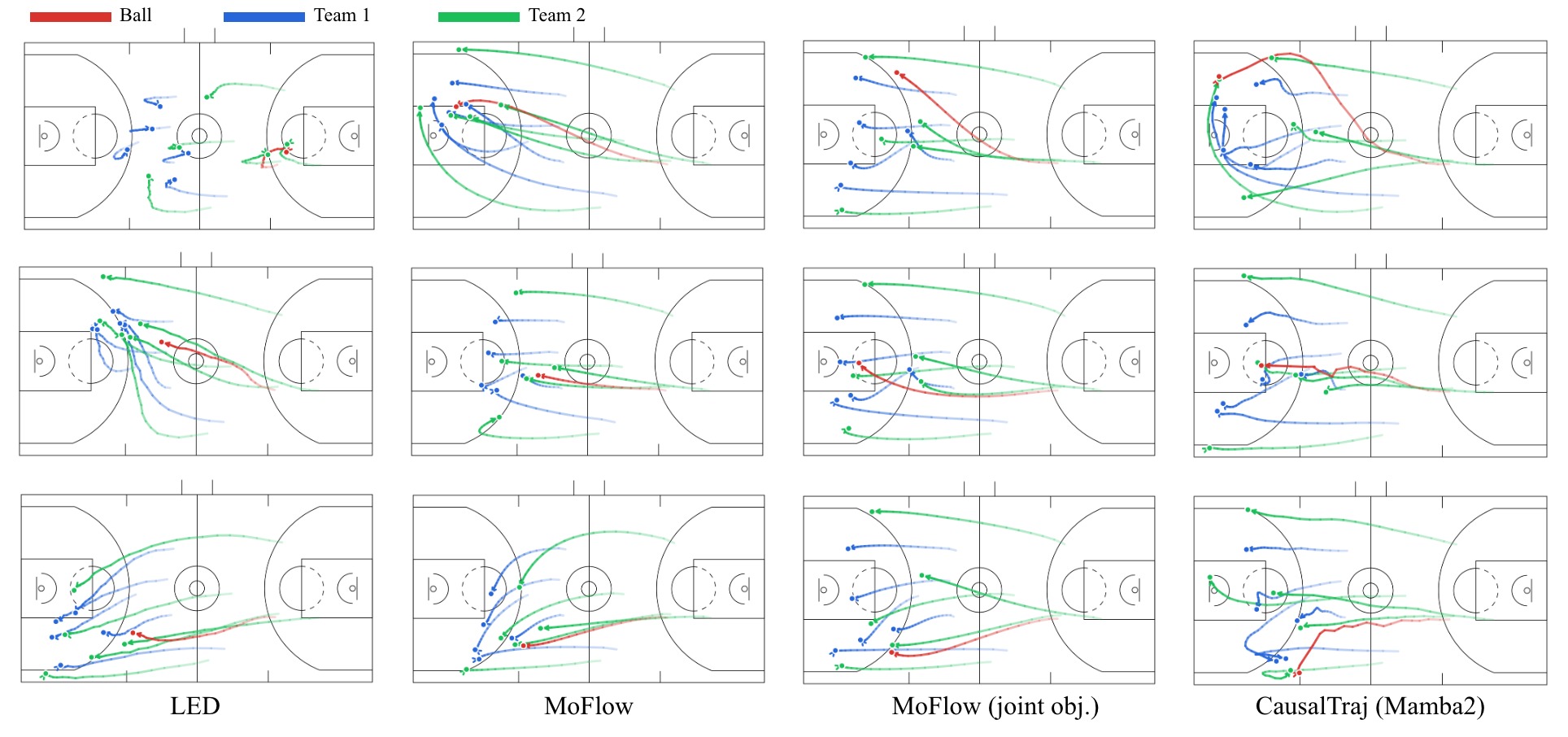} 
\caption{3 sampled scenarios (based on the same historical context) generated from each model.}
\label{fig:samples}
\end{figure*}

Figure~\ref{fig:samples} shows multiple future scenarios generated by each model given the same observed history. For LED and the default MoFlow, trajectories within each sampled scenario appear relatively homogeneous across agents (travels towards same point/direction). Since these models were not trained with an explicit joint objective, the agent's predictions at the same output branch index appear to correspond more strongly to a marginal mode rather than a mode of coordinated multi-agent outcome. On the other hand MoFlow trained with joint objective produces more coherent scenarios, where players present more strategic and coordinated positional allocation. Compared to MoFlow (joint obj.), CausalTraj further yields more diverse and physically plausible behaviors, e.g. players change directions abruptly at times to track others, ball passes more often travel faster and along straight paths rather than unrealistic arcs. Additional animations are available on our project page\footnote{\url{https://causaltraj.github.io}}
. 

Despite improved scenario coherence and accuracy with CausalTraj, occasional implausible behaviors remain (e.g., a player carrying the ball with an unrealistically large gap between them; ball collision with court boundary). We conjecture this is partly due to the limited player-ball covariance learning capacity afforded in the model. Future work will explore general, distribution-driven approaches to improve traversal dynamics modeling.

\subsection{Ablation}
We conduct further analyses to identify key components contributing to our model’s performance. The results are shown in Table~\ref{tab:bballu_ablation}.
First, we replace the \textit{Spatial Relation Transformer Encoder} (SRTE) with a standard Transformer encoder of comparable parameter count. This results in slightly degraded performance on the joint metrics (\textit{minJADE}, \textit{minJFDE}), confirming that explicitly encoding spatial relationships among agents provides measurable benefit for multi-agent modelling. 

Next, we study the effect of probabilistic modeling capacity. 
Reducing the mixture of eight Gaussian components to a single Gaussian noticeably degrades performance, as does restricting inference to sampling only from the mixture component means. 
These observations highlight the importance of modeling multimodality and spatial covariance structure for capturing the complex motion patterns present in sports trajectory data.

\begin{table}[t]
\centering
\small
\setlength{\tabcolsep}{4pt}
\begin{tabular}{lcccc}
\toprule
& \shortstack[c]{\textbf{CausalTraj}\\\textbf{(Mamba2)}} 
& \shortstack[c]{\textbf{No SRTE}} 
& \shortstack[c]{\textbf{Single}\\\textbf{Gaussian}} 
& \shortstack[c]{\textbf{Comp. Mean}\\\textbf{Sampling}} \\
\midrule
\textit{minJADE$_{20}$} & 0.97 & 0.99 & 1.03 & 1.05 \\
\textit{minJFDE$_{20}$} & 1.77 & 1.81 & 1.86 & 2.13 \\
\bottomrule
\end{tabular}
\caption{Ablation on the Basketball-U dataset (joint metrics, 20-frame horizon). Lower is better.}
\label{tab:bballu_ablation}
\end{table}

\section{Conclusion}
In this work, we introduced CausalTraj, a model for joint multi-agent trajectory forecasting in team sports. By modeling temporal causality and spatial dependencies among agents, CausalTraj demonstrated improvement in capturing the strategic interactions and collective dynamics
that underlie coordinated gameplay.

Across multiple team sports benchmark datasets, CausalTraj achieves competitive per-agent predictive accuracy compared to state-of-the-art baselines and substantially improves joint accuracy, while producing diverse, plausible, team-consistent trajectory scenarios. Beyond performance improvement, our approach contributes a practical baseline for studying coordinated behaviors in multi-agent sport settings in the future.

Future work will explore distributionally grounded methods to improve joint modelling and extend CausalTraj toward controllable, conditional generation of tactical patterns. We also plan to investigate improved metrics for evaluating multi-agent realism, providing a more robust link between quantitative assessment and perceptual coherence. We hope this work encourages broader adoption of joint metrics and coherence-driven modeling in multi-agent sports analytics and related domains.

\section{Acknowledgements}
I thank my wife Joyce for inspiring me to find the courage to pursue what I have always aspired to, and for her unwavering belief in me throughout my independent research journey.

\bibliography{aaai2026}


\appendix
\section{Training Details}
\label{appendix:training}
\subsection{Loss Function Derivation}
Each agent's $2\times2$ covariance block $\hat{\Sigma}_{t,m,n}$ is
parameterized by a lower-triangular Cholesky factor:
\[
L_{t,m,n} =
\begin{bmatrix}
\exp(\ell^{(11)}_{t,m,n}) & 0\\
\ell^{(21)}_{t,m,n} & \exp(\ell^{(22)}_{t,m,n})
\end{bmatrix}\]
\[
\hat{\Sigma}_{t,m,n} = L_{t,m,n}L_{t,m,n}^{\top}.
\]
This guarantees positive definiteness while allowing unconstrained
learning of the log-scale and correlation terms.
During training, $\ell^{(11)}$ and $\ell^{(22)}$ are clamped before exponentiation for numerical stability.

At each timestep $t$, the model actually predicts unnormalized mixture
logits $\log \hat{\pi}_{t,m}$ together with component parameters
$\{\hat{\mu}_{t,m}, L_{t,m}\}_{m=1}^{M}$.
The Gaussian log-density factorizes across agents as:
{\small
\begin{align}
\log \mathcal{N}(Y_t;\hat{\mu}_{t,m},\hat{\Sigma}_{t,m})
&= -\tfrac{1}{2}\!
\sum_{n=1}^{N}
\Big[
\|L_{t,m,n}^{-1}(Y_{n,t}-\hat{\mu}_{t,m,n})\|_2^2
\nonumber\\[-2pt]
&\quad +\, 2\log\det L_{t,m,n}
+ 2\log(2\pi)
\Big].
\end{align}
}

Instead of explicitly normalizing mixture weights,
we compute the mixture log-likelihood directly in log-space
using the numerically stable log-sum-exp form:
{\small
\begin{align}
\log p(Y_t)
&= \operatorname*{logsumexp}_{m}
\Big(
\log \hat{\pi}_{t,m}
+ \log \mathcal{N}(Y_t;\hat{\mu}_{t,m},\hat{\Sigma}_{t,m})
\Big)
\nonumber\\[-2pt]
&\quad -\,
\operatorname*{logsumexp}_{m}(\log \hat{\pi}_{t,m}).
\end{align}
}

The per-timestep negative log-likelihood is then
$\mathcal{L}_{\text{NLL}} = -\log p(Y_t)$.

This formulation allows stable gradient propagation through both the
mixture logits and covariance parameters.
All logarithms and matrix operations are computed in double precision
to avoid numerical instability.

\subsection{Optimization}
We train using the AdamW optimizer~\cite{DBLP:journals/corr/abs-1711-05101} with a OneCycle learning-rate schedule~\cite{DBLP:journals/corr/abs-1708-07120}. 
The maximum learning rate is set to $0.02$, and the weight decay to $0.01$.

\section{Model Details}
\label{appendix:model}
Model sizes for NBA dataset: 
\begin{itemize}
\item CausalTraj (Causal PointNet): 3.0M params
\item CausalTraj (Mamba2): 3.2M params
\end{itemize}

The model hyperparameters are detailed in Table~\ref{tab:model_config}.

\begin{table}[!t]
\centering
\small
\begin{tabular}{l l}
\toprule
\textbf{Component} & \textbf{Hyperparameters} \\
\midrule
\multicolumn{2}{l}{\textbf{Agent Embedding}} \\
\hspace{1em} Dim & 64 \\[3pt]

\multicolumn{2}{l}{\textbf{Causal PointNet Encoder} (*choice)} \\
\hspace{1em} $d_{\text{hidden}}$ & 64 \\
\hspace{1em} MLP depths & [1, 2, 2] \\[3pt]

\multicolumn{2}{l}{\textbf{Mamba2 Encoder} (*choice)} \\
\hspace{1em} Projector MLP depth & 2 \\
\hspace{1em} $n_{\text{layer}}$ & 3 \\
\hspace{1em} $d_{\text{model}}$ & 64 \\
\hspace{1em} $d_{\text{state}}$ & 128 \\
\hspace{1em} $d_{\text{conv}}$ & 4 \\
\hspace{1em} Expansion factor & 4 \\
\hspace{1em} Head dim & 16 \\
\hspace{1em} Groups & 1 \\
\hspace{1em} Chunk size & 32 \\
\hspace{1em} Bias / conv bias & False / False \\[3pt]

\multicolumn{2}{l}{\textbf{Standard Inter-Agent Transformer Encoder}} \\
\hspace{1em} Num blocks & 4 \\
\hspace{1em} $d_{\text{model}}$ & 128 \\
\hspace{1em} $n_{\text{head}}$ & 8 \\
\hspace{1em} $d_{\text{ff}}$ & 512 \\[3pt]

\multicolumn{2}{l}{\textbf{Spatial Relation Transformer Encoder}} \\
\hspace{1em} Num blocks & 4 \\
\hspace{1em} $d_{\text{model}}$ & 128 \\
\hspace{1em} $n_{\text{head}}$ & 8 \\
\hspace{1em} $d_{\text{ff}}$ & 256 \\[3pt]

\multicolumn{2}{l}{\textbf{Scene Agg \& Prediction Heads}} \\
\hspace{1em} Agentwise MLP depth & 1 \\
\hspace{1em} Agentwise MLP dim & 64 \\
\hspace{1em} Scene agg MLP depth & 3 \\
\hspace{1em} Scene agg MLP dim & 768 \\
\hspace{1em} Prediction head dim & 448 \\ [3pt]

\bottomrule
\end{tabular}
\caption{Default CausalTraj model configuration for the NBA dataset.}
\label{tab:model_config}
\end{table}

\end{document}